\documentclass[letterpaper, 10 pt, conference]{ieeeconf}  

\IEEEoverridecommandlockouts                              

\pdfoutput=1

\usepackage{graphicx}
\usepackage{amssymb}  

\usepackage{url}
\usepackage{xcolor}
\usepackage[citecolor=red,bookmarksopen,bookmarksnumbered,urlcolor=red]{hyperref}

\usepackage{xifthen}
\usepackage{xparse}
\usepackage{subdepth}
\usepackage{leftidx}
\usepackage{bm}
\usepackage{etoolbox}
\usepackage{amsmath}
\usepackage{multirow, makecell}

\newcommand{\bbm}{\begin{bmatrix}}
\newcommand{\ebm}{\end{bmatrix}}

\DeclareMathAlphabet{\mybf}{OT1}{ptm}{b}{n} 
\newcommand{\mybs}[1]{{\bm{#1}}} 
\DeclareMathAlphabet{\mybfi}{OML}{cmm}{b}{it}

\newcommand{\mbf}[1]{
\ifcat\noexpand#1\relax 
\mybs{#1}
\else
\mybf{#1}
\fi
}

\newcommand{\mbfbar}[1]{{\overline{\mbf{#1}}}}
\newcommand{\mbfhat}[1]{{\hat{\mbf{#1}}}}
\newcommand{\mbftilde}[1]{{\tilde{\mbf{#1}}}}

\newcommand{\mbfdot}[1]{{\dot {\mbf{#1}}}}

\NewDocumentCommand{\mbfidentity}{o}{\IfValueTF{#1}{\mbf{I}_{#1\hspace{\rightshift}}}{\mbf{I}}}
\NewDocumentCommand{\mbfzero}{oo}{\IfValueTF{#1}{\mbf{0}_{#1\times#2\hspace{\rightshift}}}{\mbf{0}}}

\newcommand{\cframe}[1]{{\smash{\protect\underrightarrow{\mathcal{F}}_{#1}}}}

\newcommand{\homo}[1]{{\mybfi{#1}}}

\newcommand{\mbfh}[1]{{\homo{#1}}}

\newcommand{\crossmx}[1]{\left[{#1}\right]^\times}

\newlength{\leftshift}
\newlength{\rightshift}
\newcommand{\pos}[2]{\leftidx{_{#1}}{ \mbf r}{_{#2\hspace{\rightshift}}}} 
\newcommand{\posbar}[2]{\leftidx{_{#1}}{\mbfbar r}{_{#2\hspace{\rightshift}}}} 
\newcommand{\poshat}[2]{\leftidx{_{#1}}{\mbfhat r}{_{#2\hspace{\rightshift}}}} 

\NewDocumentCommand{\vel}{moo}{
	\IfValueTF{#1}{\leftidx{_{#1}}}{}{\mbf v}{\IfValueTF{#2}{_{#2#3\hspace{\rightshift}}}{}}}
\NewDocumentCommand{\veltilde}{moo}{
	\IfValueTF{#1}{\leftidx{_{#1}}}{}{\mbftilde v}{\IfValueTF{#2}{_{#2#3\hspace{\rightshift}}}{}}}
\NewDocumentCommand{\velbar}{moo}{
	\IfValueTF{#1}{\leftidx{_{#1}}}{}{\mbfbar v}{\IfValueTF{#2}{_{#2#3\hspace{\rightshift}}}{}}}
\NewDocumentCommand{\velhat}{moo}{
	\IfValueTF{#1}{\leftidx{_{#1}}}{}{\mbfhat v}{\IfValueTF{#2}{_{#2#3\hspace{\rightshift}}}{}}}
\NewDocumentCommand{\veldot}{moo}{
	\IfValueTF{#1}{\leftidx{_{#1}}}{}{\mbfdot v}{\IfValueTF{#2}{_{#2#3\hspace{\rightshift}}}{}}}

\NewDocumentCommand{\acc}{moo}{
	\IfValueTF{#1}{\leftidx{_{#1}}}{}{\mbf a}{\IfValueTF{#2}{_{#2#3\hspace{\rightshift}}}{}}}
\NewDocumentCommand{\acctilde}{moo}{
	\IfValueTF{#1}{\leftidx{_{#1}}}{}{\mbftilde a}{\IfValueTF{#2}{_{#2#3\hspace{\rightshift}}}{}}}
\NewDocumentCommand{\accbar}{moo}{
	\IfValueTF{#1}{\leftidx{_{#1}}}{}{\mbfbar a}{\IfValueTF{#2}{_{#2#3\hspace{\rightshift}}}{}}}
\NewDocumentCommand{\acchat}{moo}{
	\IfValueTF{#1}{\leftidx{_{#1}}}{}{\mbfhat a}{\IfValueTF{#2}{_{#2#3\hspace{\rightshift}}}{}}}
\NewDocumentCommand{\accdot}{moo}{
	\IfValueTF{#1}{\leftidx{_{#1}}}{}{\mbfdot a}{\IfValueTF{#2}{_{#2#3\hspace{\rightshift}}}{}}}

\NewDocumentCommand{\rotvel}{moo}{
	\IfValueTF{#1}{\leftidx{_{#1}}}{}{\mbf \omega}{\IfValueTF{#2}{_{#2#3\hspace{\rightshift}}}{}}}
\NewDocumentCommand{\rotveltilde}{moo}{
	\IfValueTF{#1}{\leftidx{_{#1}}}{}{\mbftilde \omega}{\IfValueTF{#2}{_{#2#3\hspace{\rightshift}}}{}}}
\NewDocumentCommand{\rotvelbar}{moo}{
	\IfValueTF{#1}{\leftidx{_{#1}}}{}{\mbfbar \omega}{\IfValueTF{#2}{_{#2#3\hspace{\rightshift}}}{}}}
\NewDocumentCommand{\rotvelhat}{moo}{
	\IfValueTF{#1}{\leftidx{_{#1}}}{}{\mbfhat \omega}{\IfValueTF{#2}{_{#2#3\hspace{\rightshift}}}{}}}
\NewDocumentCommand{\rotveldot}{moo}{
	\IfValueTF{#1}{\leftidx{_{#1}}}{}{\mbfdot \omega}{\IfValueTF{#2}{_{#2#3\hspace{\rightshift}}}{}}}

\newcommand{\C}[2]{ {\mbf C}   {_{#1#2\hspace{\rightshift}} }     } 
\newcommand{\Cbar}[2]{{\mbfbar C}{_{#1#2\hspace{\rightshift}}}} 
\newcommand{\T}[2]{{\mbfh T}{_{#1#2\hspace{\rightshift}}}} 
\newcommand{\q}[2]{{\mbf q}{_{#1#2\hspace{\rightshift}}}} 

\newcommand{\gravity}[1]{\leftidx{_{#1}}{\mbf g}} 
\newcommand{\Exp}[1]{\text{Exp}\left( #1 \right)}
\newcommand{\attime}[1]{^{ #1 }}

\title{\LARGE \bf
Visual-Inertial SLAM with Tightly-Coupled\\Dropout-Tolerant GPS Fusion
}

\author{Simon Boche$^{1}$, Xingxing Zuo$^{1,\dagger}$, Simon Schaefer$^{1}$, Stefan Leutenegger$^{1,2}$ $^{*}$
\thanks{*This work was supported by the Technical University of Munich and Leica Geosystems AG.}
\thanks{$^{1}$Smart Robotics Lab, Department of Informatics,
Technical University of Munich, Germany.
        {\tt\small firstname.surname@tum.de}}%
\thanks{$^{2}$ Department of Computing, Imperial College London, UK}%
\thanks{$^\dagger$ Corresponding author (Email: {\tt\small xingxing.zuo@tum.de}).}
}

\begin{document}

\maketitle
\thispagestyle{empty}
\pagestyle{empty}

\begin{abstract}
Robotic applications are continuously striving towards higher levels of autonomy. To achieve that goal, a highly robust and accurate state estimation is indispensable. Combining visual and inertial sensor modalities has proven to yield accurate and locally consistent results in short-term applications. Unfortunately, visual-inertial state estimators suffer from the accumulation of drift for long-term trajectories. To eliminate this drift, global measurements can be fused into the state estimation pipeline. The most known and widely available source of global measurements is the Global Positioning System (GPS). In this paper, we propose a novel approach that fully combines stereo Visual-Inertial Simultaneous Localisation and Mapping (SLAM), including visual loop closures, with the fusion of global sensor modalities in a tightly-coupled and optimisation-based framework. Incorporating measurement uncertainties, we provide a robust criterion to solve the global reference frame initialisation problem. Furthermore, we propose a loop-closure-like optimisation scheme to compensate drift accumulated during outages in receiving GPS signals.
Experimental validation on datasets and in a real-world experiment demonstrates the robustness of our approach to GPS dropouts as well as its capability to estimate highly accurate and globally consistent trajectories compared to existing state-of-the-art methods.
\end{abstract}

\section{Introduction}
\label{sec:introduction}
Robust, accurate and globally consistent state estimation is a fundamental requirement for mobile robotic applications on the road to full autonomy.
Sensor fusion is at the core of most common state estimation algorithms. Recently, the combination of visual and inertial measurements provided by cameras and Inertial Measurement Units (IMUs) has proven to provide high accuracy and robustness. Exemplary for state-of-the-art Visual-Inertial Odometry (VIO) or Visual-Inertial SLAM (VI-SLAM) approaches that have shown impressive results, we name MSCKF~\cite{MSCKF}, BASALT~\cite{basalt}, ORB-SLAM3~\cite{orbslam3}, Kimera~\cite{kimera}, VINS-Fusion~\cite{VINS-Fusion} and the evolved OKVIS2~\cite{okvis2}. All of the aforementioned approaches have shown that visual-inertial measurements enable accurate and locally consistent pose estimation at high update rates. However, despite loop closure optimisation in VI-SLAM systems, visual-inertial approaches still suffer from the fact that they are accompanied with significant accumulated drift for long-term trajectories. This is largely caused by sensor noise, calibration errors,  modeling approximations in the estimator design, and especially, the lack of global measurements. While some applications will function well despite such significant drifts, we should strive to minimise these effects for applications where they matter, but also have to acknowledge that some extent of drift will be unavoidable in the absence of global measurements. 
Thankfully, many scenarios offer availability of global measurements, which do not depend on the distance travelled, and can be fused into state estimation. In~\cite{VINS-Fusion}, the authors introduced a general framework to feed VI-SLAM with additional global sensor modalities such as magnetometer and barometer measurements, and, most importantly, GPS measurements. Several approaches have been proposed in the field of fusing global measurements with visual-inertial navigation to eliminate the drift.

In this work, we propose a GPS-fused VIO and VI-SLAM system, which incorporate global position factors in an optimisation-based visual-inertial framework~\cite{okvis2}. In particular, we point out the following key contributions of this paper: 
\begin{itemize}
    \item We propose a novel VI-SLAM system that fuses stereo visual-inertial state estimation and global position measurements together with visual loop closure optimisation in a tightly-coupled way. It tries to use the best of global measurements to alleviate the drift of the pose estimation.
    \item We propose an uncertainty-aware method to initialise and estimate the 4 degrees-of-freedom (DoF) extrinsic transformation (yaw + translation) between a global reference frame and the VIO world reference frame, in order to fuse the global measurements into the VI estimator. Once the extrinsics are well-estimated, the global reference frame will be fixed, which simplifies the state estimation complexity and hence increases the ease of use.
    \item To cope with inevitable GPS signal outages, a re-initialisation of the GPS-VIO extrinsics will be triggered once measurements are received again after long-period GPS dropouts. In that case, the re-initialisation provides an estimate of the accumulated drift during the GPS signal outage. In a loop-closure like manner, this drift can be compensated by a global alignment approach, that uses rotation averaging and position error distribution to yield globally consistent trajectories.
\end{itemize}
The proposed approach is evaluated quantitatively and qualitatively on well-known publicly-available datasets as well as in a real-world experiment on self-recorded data. Moreover, we perform a study on the robustness of our approach in case of different outage patterns.

The remainder of this paper is structured as follows: in Section~\ref{sec:related-work}, we provide a brief overview of related literature in the field of GPS-aided VIO and VI-SLAM. Section~\ref{sec:preliminaries} gives an introduction into the notation and key concepts used throughout this work. Based on that, we will present our approach for GPS-aided VI estimation in Section~\ref{sec:method}. Following that, in Section~\ref{sec:evaluation}, we demonstrate its performance in an experimental evaluation. Finally, Section~\ref{sec:conclusions} concludes this paper and gives an outlook on future research directions. 
\section{Related Work}
\label{sec:related-work}
Visual-Inertial Navigation Systems (VINS) are often classified into two groups of algorithms: \textit{filter-based} and \textit{optimisation-based} methods. This classification naturally also applies to GPS-aided approaches. Filter-based methods estimate the probability distribution of the states including poses and landmarks~\cite{MSCKF,lynen2013robust}. Optimisation-based approaches formulate a non-linear least-squares minimisation problem which is solved by bundle-adjustment~\cite{okvis2,lynen2013robust}. From the perspective of fusing measurements, we can distinguish between \textit{tightly-coupled} and \textit{loosely-coupled} methods.
Loosely-coupled approaches estimate states separately for different sensors, and fuse the estimation results from different sensors afterwards. Conversely, tightly-coupled methods jointly leverage the measurements from different sensor modalities for state estimation.  

Early work on fusing global sensors is dominated by filter-based methods. A very large number of these methods built upon the seminal work in~\cite{MSCKF} where the authors propose an Extended Kalman Filter (EKF) to tackle real-time visual-inertial navigation.
\cite{lynen2013robust} and~\cite{shen2014multi} use an EKF and an Unscented Kalman Filter (UKF), respectively, to fuse different sensor modalities, such as LiDAR and GPS measurements. 
A recent approach for GPS-aided state estimation is~\cite{lee2020intermittent}, which additionally estimates the IMU-GPS spatial extrinsics as well as the sensor time offset online in an EKF-based MSCKF~\cite{MSCKF} framework. 
Filter-based methods are in general computationally efficient. Nevertheless, as investigated in~\cite{strasdat2010real}, optimisation-based approaches potentially deliver superior results compared to filter-based approaches.

Regarding the realm of optimization-based methods, VINS-Fusion~\cite{VINS-Fusion} and GOMSF~\cite{GOMSF} fuse global position measurements with pose estimates from an independent VIO in the back-end pose-graph optimisation. Both of these methods are loosely-coupled methods to fuse global sensor measurements. As opposed to  VINS-Fusion~\cite{VINS-Fusion}, GOMSF~\cite{GOMSF} introduces an additional virtual node in the pose-graph to constrain the absolute orientation of the local frame in a global reference frame. \cite{yu2019gps} constitutes another example of a loosely-coupled approach fusing the GPS measurements in a refinement step for a flexible number of cameras. As loosely-coupled approaches do not fully exploit the available information given by different sensor modalities, they are limited in terms of accuracy compared to tightly-coupled approaches. To achieve higher accuracy in pose estimation, it is desirable to fuse GPS measurements with the VI-SLAM system in a tightly-coupled way.
Cioffi et al.~\cite{cioffi2020tightly} propose a tightly-coupled approach fusing GPS measurements in the optimisation window as global factors leveraging IMU pre-integration. Recent works in~\cite{GVINS} and~\cite{liu2020optimization} have also proposed tightly-coupled optimization-based frameworks which consider not only global position measurements but also pseudo range and Doppler shift errors.

While~\cite{cioffi2020tightly} assumes global position measurements given in the visual-inertial reference frame, in practice, global measurements will usually be given in an independent global reference frame. A 4 DoF transformation has to be estimated to align the two frames. Initialisation of this global frame alignment problem has been addressed by several means in previous works. In the loosely-coupled~\cite{GOMSF}, this extrinsic transformation is initialised based on SVD leveraging correspondences of local and global position measurements. \cite{lee2020intermittent},~\cite{GVINS} and~\cite{liu2020optimization} estimate the GPS-VIO extrinsics in a tightly-coupled fashion minimizing different cost functions. \cite{lee2020intermittent} furthermore introduces a heuristic criterion on the observability of the 4 DoF transformation based on the distance travelled. As in~\cite{GOMSF}, we also use an SVD-based initialisation of the GPS-VIO extrinsics. However, it will be incorporated in a tightly-coupled way. In contrast to the aforementioned approaches, to simplify the problem, the global reference frame will be fixed in our approach as soon as it becomes observable. Instead of applying a heuristic threshold as in~\cite{lee2020intermittent}, we introduce a fully uncertainty-aware criterion to determine the covariance of the estimated GPS extrinsics.

As a novelty, we also propose a loop closure-like optimisation approach to effectively tackle the issue of drift during longer periods of GPS outage.
\section{Preliminaries}
\label{sec:preliminaries}
\subsection{Notation and Definitions}
The classic VI-SLAM problem formulation includes several different coordinate frames. A moving body is tracked with respect to a fixed world reference frame $\cframe{W}$. Additionally, each of the sensors has a different frame. We will denote the IMU frame by $\cframe{S}$ and the camera coordinate frames by $\cframe{C_{i}}$ for $ i = 1\dots N$ cameras. Fusing GPS measurements requires to introduce a global reference frame $\cframe{G}$ which is a gravity-aligned East-North-UP (ENU) local Cartesian frame. In the world frame, gravity is indicated by $\gravity{W}$.

The rigid body transformation $\T{A}{B} \in SE(3)$ transforms homogeneous points between two frames: $\pos{A}{P} = \T{A}{B}\pos{B}{P}$, where $\pos{A}{P}$ is the position of a point $P$ in frame $\cframe{A}$. The rotational part of $\T{A}{B}$ is expressed by $\mathbf{C}_{AB} \in SO(3)$ and $\pos{A}{B}$ denotes the translation component. We also denote the rotation $\mathbf{C}_{AB}$ with its unit quaternion form $\q{A}{B}$.

$\crossmx{\cdot}$ represents the skew-symmetric matrix of a 3D vector and $\Exp{\cdot}$ stands for the exponential map of $SO(3)$ with 3D rotation vector input as used in~\cite{forster2016manifold}. Superscript indices $\left( \cdot \right)\attime{i}$ are used to represent a quantity at a time step $i$.
\subsection{Extension of OKVIS2}
In this work, we implement the proposed GPS-fused estimator based on the existing VIO/VI-SLAM system OKVIS2. In the following, we provide only a brief recap of key components related to this work. For more details of OKVIS2, please refer to~\cite{okvis2}.
The state representation in this work is:
\begin{equation}
\label{eq:state-vector}
    \mathbf{x} = \left[ \pos{W}{S}^{T} , \q{W}{S}^{T}, \vel{W}[][]^{T}, \mathbf{b}_\mathrm{g}^{T} , \mathbf{b}_\mathrm{a}^{T} \right]^{T} ,
\end{equation}
where $\pos{W}{S}$ and $\q{W}{S}$ denote the position and orientation of the IMU sensor frame in the fixed world frame, and $\vel{W}[][]$ describes the velocity of the IMU with respect to the world frame. $\mathbf{b}_\mathrm{g}$ and $\mathbf{b}_\mathrm{a}$ stand for gyroscope and accelerometer biases, respectively. In order to fuse the visual-inertial system and global measurements, we also estimate online the extrinsic transformation $ \T{G}{W} = \{ \pos{G}{W} , \q{G}{W} \}$ between the world reference frame of the visual-inertial estimator $\cframe{W}$ and the GPS reference frame $\cframe{G}$.

Global position residuals can be fused into the existing framework in a tightly-coupled way by adding them to the original non-linear least-squares cost $c_{\mathrm{VI}}\left( \mathbf{x} \right)$ in~\cite{okvis2}, consisting of visual, inertial and relative pose error residuals. The overall minimisation objective becomes
\begin{equation}
    \label{eq:overall_min_problem}
    c\left( \mathbf{x} , \T{G}{W} \right) = c_{\mathrm{VI}}\left( \mathbf{x} \right) + \frac{1}{2} \sum_{j \in \mathcal{G}} {\mathbf{e}_\mathrm{g}\attime{j}}^{T} \mathbf{W}_\mathrm{g}\attime{j} {\mathbf{e}_\mathrm{g}\attime{j}}.
\end{equation}
Here, $\mathbf{e}_\mathrm{g}\attime{j}$ is the global position error for a measurement at time step $j$ and $\mathbf{W}_\mathrm{g}\attime{j}$ is the weight matrix denoted by the inverse of the measurement covariance matrix $\boldsymbol{\Sigma}_\mathrm{g}\attime{j}$. $\mathcal{G}$ is the set of all GPS measurements that have been received so far. The global position residual as well as the corresponding covariance matrix will be derived in the following section.

OKVIS2 runs two different optimisation threads. One thread executes the real-time graph optimisation of the most recent keyframes. The full graph optimisation is launched on a secondary thread upon loop closure. The global position residuals are added as factors to both of the aforementioned graphs.
\section{Fusion of Global Position Measurements}
\label{sec:method}
In this section, the global position residuals used in~\eqref{eq:overall_min_problem} are derived. Furthermore, an initialisation strategy for determining the orientation and position of the GPS reference frame is provided, and we will explain our approach of loop closure-like optimisation after GPS dropouts.
\subsection{IMU Pre-Integration}
\label{sec:pre-int}
As camera frames and GPS measurements are received in an asynchronous fashion, we want to leverage integrated IMU measurements to deal with the spatial offset due to camera movement. To avoid repeated integration of IMU measurements whenever the estimated pose of a state changes, we will apply a modified version of the IMU pre-integration approach presented in~\cite{forster2016manifold}. Same as in the IMU error terms used by \cite{okvis2}, instead of Euler integration, a trapezoidal integration scheme is applied. Assuming IMU measurements given at time steps $k$ in the interval $\left[i , j \right]$ for a camera frame at time step $i$ and a GPS measurement at time step $j$, the following holds:
\begin{equation}
\label{eq:pre-integration:step}
    \begin{aligned}
        \mathbf{C}_{WS}\attime{j} &= \mathbf{C}_{WS}\attime{i} \boldsymbol{\alpha}_{i}^{j}, \\
        \vel{W}[][]\attime{j} &= \vel{W}[][]\attime{i} + \mathbf{C}_{WS}\attime{i}\boldsymbol{\beta}_{i}^{j} +\gravity{W}\Delta t\attime{ij}, \\
        \pos{W}{S}\attime{j} &= \pos{W}{S}\attime{i} + \vel{W}[][]\attime{i}\Delta t\attime{ij} + \frac{1}{2} \gravity{W}\Delta {t\attime{ij}}^{2} +  \mathbf{C}_{WS}\attime{i} \boldsymbol{\gamma}_{i}^{j},
    \end{aligned}
\end{equation}
where $\boldsymbol{\alpha}_{i}^{j}$, $\boldsymbol{\beta}_{i}^{j}$ and $\boldsymbol{\gamma}_{i}^{j}$ are pre-integration terms which only depend on the IMU measurements between the two frames. These terms can be computed as
\begin{equation}
\label{eq:pre-integration:terms}
    \begin{aligned}
        \boldsymbol{\alpha}_{i}^{j} &= \prod_{k=i}^{j-1} \Exp{ \left( \rotvelbar{S}[][]\attime{k} - \mathbf{b}_{g} \right) \Delta t} , \\
        \boldsymbol{\beta}_{i}^{j} &= \sum_{k=i}^{j-1} \frac{1}{2} \left( \boldsymbol{\alpha}_{i}^{k} + \boldsymbol{\alpha}_{i}^{k+1} \right) \left( \accbar{S}[][]\attime{k} - \mathbf{b}_{a}\right) \Delta t , \\
        \boldsymbol{\gamma}_{i}^{j} &= \sum_{k=i}^{j-1} \boldsymbol{\beta}_{i}^{k} \Delta t +  \frac{1}{4} \left( \boldsymbol{\alpha}_{i}^{k} + \boldsymbol{\alpha}_{i}^{k+1} \right) \left( \accbar{S}[][]\attime{k} - \mathbf{b}_{a}\right) \Delta t^{2}.
    \end{aligned}
\end{equation}
Note that so far, it is assumed that the gyroscope and accelerometer biases $\mathbf{b}_{g}$ and $\mathbf{b}_{a}$ are constant in the integration interval. If the bias estimate changes, we apply a bias update by linearisation as in~\cite{forster2016manifold}. In~\eqref{eq:pre-integration:terms}, $\rotvelbar{S}[][]\attime{k}$ is the mean of two subsequent gyro measurements $\rotveltilde{S}[][]\attime{k}$, $\rotveltilde{S}[][]\attime{k+1}$. The same applies to accelerometer measurements $\accbar{S}[][]\attime{k}$.

The pre-integration in Eqs.~\eqref{eq:pre-integration:step} allows us to express the state at IMU time step $j$ as a function of the estimated states at time step $i$, i.e.\ $\mbfhat{x}^j= f(\mbf{x}^i)$.
Further note that the covariance of $\mbfhat{x}^j$ can be also computed conditioned on the given $\mbf{x}^i$ and the IMU pre-integtration covariance $\mbf{\Sigma}_{\mathrm{x}}^j$~\cite{forster2016manifold} via linear error propagation.  
%
\begin{figure}[t]
    \centering
    \includegraphics[width=0.8\linewidth]{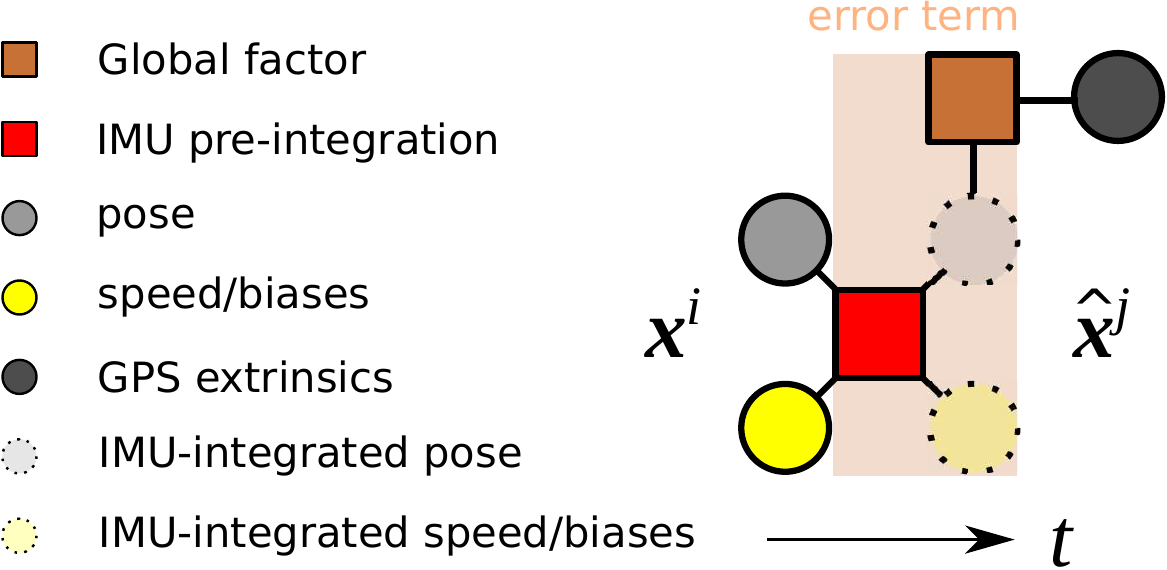}
    \caption{To evaluate the global position residual, the propagated state $\hat{\mathbf{x}}\attime{j}$ at time step $j$ of the measurement is propagated from the state vector $\mathbf{x}\attime{i}$ at time step $i$ of the camera frame by leveraging IMU pre-integration. This intermediate state is only used for evaluation of the global factor.}
    \label{fig:gpsErrorTermVis}
\end{figure}
\subsection{GPS Error Formulation}
The GPS residual for a measurement $\mathbf{z}\attime{j}$ at a time step $j$ can be formulated as:
\begin{equation}
\label{eq:gps res}
    \mathbf{e}_\mathrm{g}^{j} = \mathbf{z}\attime{j} - 
    \left[\C{G}{W} \left( \poshat{W}{S}\attime{j} + \hat{\mathbf{C}}_{WS}\attime{j} \pos{S}{A} \right) + \pos{G}{W} \right].
\end{equation}
Hereby, $\pos{S}{A}$ considers the position of the GPS antenna in the IMU sensor frame which is assumed to be known beforehand. $\poshat{W}{S}\attime{j}$ and $\hat{\mathbf{C}}_{WS}\attime{j}$ are predictions of the IMU poses at time step $j$ which can be obtained from a previous state at time step $i$ through IMU pre-integration (Eq.~\eqref{eq:pre-integration:step}~-~\eqref{eq:pre-integration:terms}). This is visualised in Fig.~\ref{fig:gpsErrorTermVis}. Compared to~\cite{cioffi2020tightly}, the formulation in~\eqref{eq:gps res} extends the global residual by also considering a 4 DoF transformation $\T{G}{W}$ between the global and the visual-inertial reference frame.

We generally use the perturbation $\delta \mbf{\chi}_T = [\delta \mbf{r}, \delta \mbf{\alpha}]$ for poses around linearisation points for translation $\mbfbar{r}$ and orientation $\mbfbar{C}$:
\begin{equation}
\label{eq:error-states}
    \begin{aligned}
    \mbf{r} &= \bar{\mbf{r}} + \delta \mbf{r}, \\
    \mbf{C} &= \Exp {\delta \boldsymbol{\alpha}} \mbfbar{C},
    \end{aligned}
\end{equation} 
where an additional subscript is used to indicate the respective transformation. With this, we further define the error state as
$\delta \boldsymbol{\chi} = [\delta \mbf{\chi}_{T_{WS}}, \delta\mbf{\chi}_\mathrm{sb}]$, with $\delta\mbf{\chi}_\mathrm{sb}$ denoting an additive perturbation of the speed and biases.

The Jacobians with respect to the pose error states of the predicted pose $\hat{\boldsymbol{T}}_{WS}$ based on the IMU measurements and the relative pose between the GPS and world frame $\boldsymbol{T}_{GW}$ are given by:
\begin{equation}
\label{eq:error-jacobians}
    \begin{aligned}
        \frac{ \partial \mathbf{e}_\mathrm{g}\attime{j} }
         { \partial \delta \hat{\boldsymbol{\chi}}_{{T}_{WS}}\attime{j}}
        &= 
        \begin{bmatrix}
            -\Cbar{G}{W}
            &
            \Cbar{G}{W} \crossmx{ \Cbar{W}{S} \pos{S}{A} }
        \end{bmatrix},
        \\
        \frac{ \partial \mathbf{e}_\mathrm{g}\attime{j} }
         { \partial \delta \boldsymbol{\chi}_{T_{GW}}}
        &= 
        \begin{bmatrix}
            -\mathbf{I}_{3\times3}
            &
            \crossmx{ \Cbar{G}{W} \left( \posbar{W}{S} + \Cbar{W}{S}\pos{S}{A}\right) }
        \end{bmatrix}.
    \end{aligned}
\end{equation}
The predicted poses $\mbfhat{x}_{{T}_{WS}}^j$ are not estimated states in the pose-graph optimisation. Since the predicted measurement is a function of the estimated states at time step $i$, i.e.\ $\mbfhat{x}^j= f(\mbf{x}^i)$ (see Eqs.~\eqref{eq:pre-integration:step}), we can actually optimise $\mbf{x}^i$ in the estimated states. The Jacobian with respect to $\mbf{x}^i$ can be computed through chain rule:
\begin{equation}
    \frac{ \partial \mathbf{e}_\mathrm{g}\attime{j} }
         { \partial \delta \boldsymbol{\chi}\attime{i}}
         =
     \frac{ \partial \mathbf{e}_\mathrm{g}\attime{j} }
         { \partial \delta \hat{\boldsymbol{\chi}}_{{T}_{WS}}\attime{j}}\hspace{1mm}
     \frac{ \partial \delta \hat{\boldsymbol{\chi}}_{{T}_{WS}}\attime{j} }
         { \partial \delta \boldsymbol{\chi}\attime{i}},
\end{equation}
where the second term can be computed from the IMU pre-integration, see \ref{sec:pre-int}.

The overall covariance matrix $\boldsymbol{\Sigma}_\mathrm{g}\attime{j}$ for GPS residual in Eq.~\eqref{eq:gps res} originates from two parts. The first one is the raw GPS measurement noises with covariance matrix $\boldsymbol{\Sigma}_\mathrm{g_0}\attime{j}$. Additionally, the noises from IMU pre-integration mesurements between the GPS measurement and the camera frame need to be considered, which is denoted by $\boldsymbol{\Sigma}_{\mathrm{x}}\attime{j}$. 
Thus, $\boldsymbol{\Sigma}_\mathrm{g}\attime{j}$ can be derived by: 
\begin{equation}
\label{eq:overall-gps-covariance}
    \boldsymbol{\Sigma}_\mathrm{g}\attime{j} = \boldsymbol{\Sigma}_\mathrm{g_0}\attime{j} + \mathbf{J} \boldsymbol{\Sigma}_{\mathrm{x}}\attime{j} \mathbf{J}^{T},
\end{equation}
where $\mathbf{J}$ denotes the Jacobian of  $\mathbf{e}_\mathrm{g}^{j} $ with respect to $\hat{\boldsymbol{\chi}}_{{T}_{WS}}\attime{j}$.
%
%
\subsection{Global Reference Frame Initialisation}
\label{sec:method_init}
For the later proposed global alignment strategy, a reliable initialisation of the global reference frame, $\cframe{G}$, with respect to the visual-inertial reference frame, $\cframe{W}$, is crucial. It is desirable to define an uncertainty-aware criterion for the observability of the 4-DoF extrinsic transformation between the reference frames based on the received measurements. 
In a first step, we therefore obtain an initial solution for the GPS extrinsics from correspondences of global measurements and poses in the world reference frame using the SVD-based alignment method presented in~\cite{posyaw} . 
Subsequently, we need to evaluate the reliability of this initial solution. For $N_\mathrm{g}$ received measurements, we first compute the global position errors $ \mathbf{e}_\mathrm{g}^{j}$ ($j=1,\dots,N$) as well as corresponding error Jacobians $\mathbf{E}_\mathrm{g}^{j}$ with respect to $ \delta \boldsymbol{\chi}_{T_{GW}}$ based on equations~\eqref{eq:gps res} and~\eqref{eq:error-jacobians}. Actually, we only consider the part of the orientation corresponding to the yaw angle $\theta$. 
The covariance matrix $\mathbf{P}$ for the 4 DoF transformation can be estimated as the inverse of the approximate Hessian $\mathbf{H}$:
\begin{equation}
\label{eq:init_hessian}
    \mathbf{H} = \sum_{j=1}^{N_\mathrm{g}} \left( {\mathbf{E}_\mathrm{g}^{j}}^{T} \mathbf{W}_\mathrm{g}\attime{j} \mathbf{E}_\mathrm{g}^{j} \right) .
\end{equation}
We examine the variance $p_{\theta \theta} $ corresponding to the yaw angle $\theta$ in $\mathbf{P}$, and define a decision threshold $\sigma_{\theta}^{2}$ on whether the global reference frame is observable as
\begin{equation}
\label{eq:init_threshold}
    p_{\theta \theta} < \sigma_{\theta}^{2}.
\end{equation}
\subsection{GPS Alignment}
\label{sec:method_align}
For the fusion of global position factors, one can distinguish three different stages: an \textit{initialisation stage}, a fully \textit{initialised stage} and a \textit{re-initialisation stage} which can, based upon the success of re-initialisation, trigger different trajectory alignment strategies aiming towards the elimination of accumulated drift during longer periods of GPS signal dropouts. 
\subsubsection{Initialisation Stage}\label{stage:init}
After having received the first GPS position measurements (at least two), an initialisation for the extrinsics of the global reference frame is determined based on the approach introduced in Section~\ref{sec:method_init} and provided to the real-time estimator. Global position factors are added to both graph optimisation problems. The transformation $\T{G}{W}$ will be estimated live. As soon as the 4 DoF GPS extrinsics are assumed to be observable according to the criterion in~\eqref{eq:init_threshold}, $\T{G}{W}$ will be fixed.
\subsubsection{Initialised Stage}\label{stage:initiialised}
After having finished initialisation of the global reference frame, it is feasible to fuse the incoming GPS measurements in the optimisation.
\subsubsection{Re-Initialisation Stage}\label{stage:reInit}
\begin{figure}[ht]
    \centering
    \includegraphics[width=\linewidth]{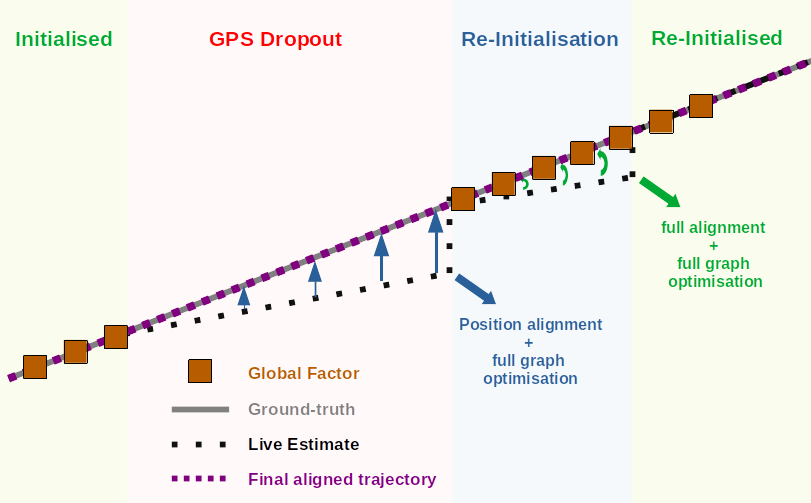}
    \caption{Conceptual visualisation of the global trajectory alignment. Due to GPS dropout, the visual inertial estimator suffers from drift in the estimate. As soon as the first GPS measurement can be received again, positions can be aligned. Upon successful re-initialisation, we can also align orientations in the graph.}
    \label{fig:alignment-vis}
\end{figure}

OKVIS2 limits the size of the optimised states and thus also the computational complexity of the problem by fixing states that are far in the past. Whenever GPS measurements are added to the graph optimisation problem, we check the fixation status of the last state that owns a global position factor. Dropouts in receiving GPS signals over a long period of time can be identified by whether the last state with a GPS factor in the graph is already fixed after the sliding-window optimisation. In that case, the trajectory might have accumulated a significant amount of drift since then. Eliminating this drift is addressed in two steps which are also conceptually visualised in Fig.~\ref{fig:alignment-vis}.
\begin{enumerate}
    \item \emph{Position Alignment}\label{posalign}
One valid GPS measurement can provide sufficient information to give a rough estimate of the drift in the position. To provide the best possible update for the live trajectory, we perform an immediate position alignment. Using linear interpolation, we distribute the positional inconsistency uniformly over the trajectory segment (equally for all states) since the last state before the loss of the GPS signal. Following the positional correction, an optimisation of the full back-end graph is started to refine the whole trajectory during GPS signal outage. 
\item \emph{Full Alignment}
In addition to the position alignment step described above, if multiple GPS measurements resume being received after outages, we will start a re-initialisation of the global reference frame, following the same approach as in the initialisation stage. From the re-initialisation process, a new estimate for the extrinsic transformation, $\T{G}{W_\mathrm{new}}$, can be obtained. Then, the discrepancy between the originally estimated $\T{G}{W}$ and the newly initialised $\T{G}{W_\mathrm{new}}$ can be calculated as:
\begin{equation}
    \T{W_\mathrm{new}}{W} = \mbfh{T}^{-1}_{GW_\mathrm{new}}\hspace{1mm}\T{G}{W},
\end{equation}
which also gives an estimate of the trajectory drift. Using this delta transformation $\T{W_\mathrm{new}}{W}$, a correction can be applied to both, the positions and the orientations of the intermediate states. Rotation averaging is used to distribute the orientation error across the states during a GPS dropout followed by another position correction. After this full alignment, an optimisation of the full graph is again triggered.
\end{enumerate}

\section{Experimental Results}
\label{sec:evaluation}
We evaluate the proposed approach both on publicly-available datasets and our own collected dataset. We conduct experiments on the well-known EuRoC~\cite{Euroc}, TUM-VI~\cite{TUMVI} and KITTI~\cite{KITTI} datasets. While KITTI provides raw GPS sensor readings in geodetic coordinates, the other two datasets provide 3D ground-truth trajectories recorded by motion capture systems. As in~\cite{cioffi2020tightly}, we simulate noisy GPS measurements for these two datasets by adding Gaussian noise to the ground-truth positions. The performance of the proposed approach will be compared quantitatively as well as qualitatively to state-of-the-art approaches.
We use Ceres~\cite{Ceres} to perform the back-end optimisation, following OKVIS2~\cite{okvis2}. The GeographicLib~\cite{GeographicLib} library is used to convert global position measurements in forms of geodetic coordinates to a local Cartesian frame with the origin at the location of the first received measurement. In the following experiments, the decision threshold for the global reference frame initialization is set to $\sigma_{\theta} = 1.0 ^\circ$ (see Eq.~\eqref{eq:init_threshold}). 
\subsection{EuRoC Dataset}
The EuRoC dataset consists of a total of eleven sequences recorded on a Micro Aerial Vehicle (MAV). The dataset provides stereo camera images, IMU measurements and ground-truth poses from a motion capture system. GPS measurements are obtained by corrupting the ground-truth positions with noises sampled from a zero-mean normal distribution $\mathcal{N} \left( \mbf{0} , \sigma_{n}^{2} \mathbf{I} \right)$. The performance of the proposed approach is assessed quantitatively by evaluating the Average Trajectory Error (ATE). Each configuration of the simulation is repeated three times and the median of the resulting ATEs is reported.
\begin{table}[t]
\caption{Median ATE [m] (final trajectory estimates) on the EuRoC Dataset Sequences. GPS noise $\sigma_{n} = 20$ cm.}
\label{tab:euroc_comparison}
\begin{center}
\begin{tabular}{c | c | c  |  c   c  || c c  }
\hline
\multirow{2}{*}{} & \multicolumn{4}{c||}{\multirow{2}{*}{VIO}} & \multicolumn{2}{c}{\multirow{2}{*}{VI-SLAM}} \\ & \multicolumn{4}{c||}{} &  \\
\hline
\multirow{2}{*}{GPS} & \multirow{2}{*}{\checkmark} & \multirow{2}{*}{\checkmark} & \multirow{2}{*}{$\times$} & \multirow{2}{*}{\checkmark}  &  \multirow{2}{*}{$\times$} & \multirow{2}{*}{\checkmark} \\ & & & & & &\\
\hline
\multirow{2}{*}{} & VINS- & TC VIO & \multicolumn{2}{ c ||}{OKVIS2} & \multicolumn{2}{c}{OKVIS2}  \\
& Fusion~\cite{VINS-Fusion} & \cite{cioffi2020tightly} & \multicolumn{2}{c ||}{(Ours)}  & \multicolumn{2}{c }{(Ours)} \\
\hline
MH01 & 0.109 & 0.022 & 0.035 & 0.019  & 0.033 & \textbf{0.018} \\
MH02 & 0.117 & 0.025 & 0.037 & 0.026 & \textbf{0.023} & 0.029 \\
MH03 & 0.223 & 0.033 & 0.076 & 0.026 & 0.030 & \textbf{0.022} \\
MH04 & 0.136 & 0.048 & 0.079 & \textbf{0.029} & 0.068 & 0.037 \\
MH05 & 0.136 & 0.039 & 0.130 & 0.034 & 0.073 & \textbf{0.032} \\
V101 & 0.088 & 0.034 & 0.050 & \textbf{0.031} & 0.033 &  \textbf{0.031} \\
V102 & 0.104 & 0.035 & 0.075 & 0.023 & 0.019 & \textbf{0.014} \\
V103 & 0.106 & 0.042 & 0.040 & 0.023 & 0.022 & \textbf{0.018} \\
V201 & 0.147 & 0.026 & 0.026 & 0.020 & 0.024 & \textbf{0.017} \\
V202 & 0.161 & 0.033 & 0.055 & 0.021 & 0.018 & \textbf{0.015} \\
V203 & 0.214 & 0.057 & 0.101 & 0.024 & 0.026 &  \textbf{0.021} \\
\hline
Avg & 0.140 & 0.036 & 0.064 & 0.025 & 0.034 & \textbf{0.023} \\
\hline
\end{tabular}
\end{center}
\end{table}
\subsubsection{Performance during full GPS availability}
We start by assuming full availability of GPS signals throughout the whole sequence. We evaluate the performance of OKVIS2 running in VIO and in VI-SLAM mode. Results are shown for the GPS-aided version as well as for the pure VI estimator. Furthermore, we compare the performance of the proposed system with two different state-of-the-art approaches, VINS-Fusion~\cite{VINS-Fusion} and~\cite{cioffi2020tightly}, with the latter denoted as tightly-coupled (TC) VIO here. For the loosely-coupled VINS-Fusion, results are generated using the open-source code with the provided default configuration files. In order to make a fair comparison to TC VIO results taken from~\cite{cioffi2020tightly} (since no code is available), isotropic noises of the GPS measurements are generated with the same standard deviation of $\sigma_{n} = 20$ cm, following~\cite{cioffi2020tightly}. Results are given in Table~\ref{tab:euroc_comparison}.

The results show that fusing global position measurements into the OKVIS2 VIO pipeline can significantly increase the accuracy. Especially for those sequences, where the VIO suffers from larger drifts, the ATE can be decreased by a large margin. 
The proposed GPS-aided VIO based on OKVIS2 clearly outperforms the loosely-coupled approach of VINS-Fusion and shows competitive to other tightly-coupled approaches like TC VIO. As the approaches are very similar, the slightly increased performance of the proposed method might be largely related to the fact that, in contrast to OKVIS2,~\cite{cioffi2020tightly} uses monocular SVO~\cite{svo} as front-end. The inclusion of visual loop closures in VI-SLAM mode leads to a minor improvement for most of the compared sequences.
\subsubsection{Robustness to GPS Dropouts}
\begin{table}[b]
\caption{Median ATE [m] of our GPS-aided VIO using different initialisation strategies on EuRoC dataset with varied GPS dropout patterns. GPS noise $\sigma_{n} = 20$ cm.}
\label{tab:euroc-dropout}
\begin{center}
\begin{tabular}{c l | c | c | c | c }
\multicolumn{2}{c |}{\multirow{2}{*}{}} & \multicolumn{4}{c}{GPS availability} \\
\multicolumn{2}{c |}{} & \multicolumn{4}{c}{(blue: on, white: off)} \\
\hline
\multirow{2}{*}{Seq.} & \multirow{2}{*}{Initialisation} & \multirow{2}{*}{\includegraphics[width=0.6cm]{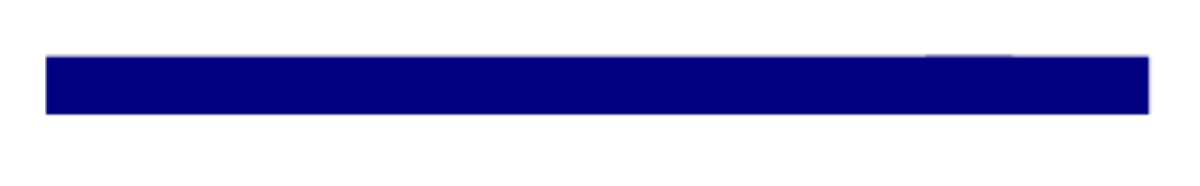}} & \multirow{2}{*}{\includegraphics[width=0.6cm]{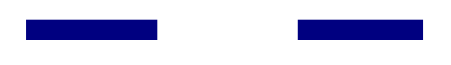}}  & \multirow{2}{*}{\includegraphics[width=0.6cm]{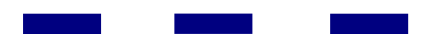}}  & \multirow{2}{*}{} \\
\multicolumn{2}{c |}{} & & & & \\
\hline
\multirow{3}{*}{MH03} & SVD once & 0.026 & 0.048 & 0.053 & \multirow{3}{*}{0.076} \\
 & SVD always & \textbf{0.025} & 0.040 & 0.041 &  \\
 & SVD crit. \& LC (Ours) & 0.026 & \textbf{0.039} & \textbf{0.038} &  \\
\hline
\multirow{3}{*}{MH04} & SVD once & 0.032 & 0.062 & 0.059 & \multirow{3}{*}{0.079} \\
 & SVD always & 0.031 & 0.063 & \textbf{0.052} &  \\
 & SVD crit. \& LC (Ours) & \textbf{0.029} & \textbf{0.039} & 0.060 &  \\
\hline
\multirow{3}{*}{MH05} & SVD once & \textbf{0.031} & 0.073 & 0.061 & \multirow{3}{*}{0.130} \\
 & SVD always & \textbf{0.031} & 0.076 & 0.068 &  \\
 & SVD crit. \& LC (Ours) & 0.034 & \textbf{0.048} & \textbf{0.056} & \\
\hline
\multirow{3}{*}{V102} & SVD once & \textbf{0.020} & 0.036 & \textbf{0.025} & \multirow{3}{*}{0.075} \\
 & SVD always & 0.022 & 0.026 & 0.026 &  \\
 & SVD crit. \& LC (Ours) & 0.023 & \textbf{0.023} & 0.028 &  \\
\hline 
\multirow{3}{*}{V203} & SVD once & 0.024 & 0.069 & \textbf{0.035} & \multirow{3}{*}{0.101} \\
 & SVD always & 0.024 & 0.065 & 0.036 &  \\
 & SVD crit. \& LC (Ours) & 0.024 & \textbf{0.045} & 0.036 &  \\
\hline
\end{tabular}
\end{center}
\end{table}
Next, we want to investigate the robustness of our approach to potential dropouts in the received GPS signals and demonstrate the benefit of the proposed initialisation method together with the global alignment strategies from sections~\ref{sec:method_init},~\ref{sec:method_align}. Therefore, we simulate different dropout patterns which are visualised next to the results in Table~\ref{tab:euroc-dropout} ($33\%$ dropout once or $20\%$ twice). Out of all EuRoC sequences, we picked the ones that suffer the most from drift. These patterns exhibit different portions of GPS signal outages, ranging from full GPS availability to total absence of GPS measurements.
As baselines, we report different strategies for the global reference frame initialisation.
In the first case, denoted as \textit{SVD once}, the global frame is initialised by SVD once in the beginning and will always use the optimised result as a starting point in the next frame, as done in the loosely-coupled GOMSF~\cite{GOMSF}. It is ensured that the distance between the first measurements used for initialisation exceeds the measurement uncertainty. The second case, \textit{SVD always}, continuously initialises $\T{G}{W}$ using SVD. In both cases the extrinsic transformation of the GPS remains variable and is always estimated live. The last case denoted by \textit{SVD crit. \& LC} refers to our proposed approach fixing the global reference frame based on measurement uncertainties and performing re-initialisation together with GPS loop closures (LC).

It can be observed that the difference between the presented approaches to initialise the global reference frame are negligible in the case of continuous GPS signal coverage. Moreover, also for shorter outages, no significant impact can be seen in the results. That changes, when it comes to longer dropouts. While the reported baseline approaches suffer from drift during the absence of global position measurements, the proposed global alignment is able to compensate for large parts of the drift and provide more robust accuracy across the different dropout patterns.
\subsection{TUM-VI Dataset}
\begin{figure}[b]
    \centering\includegraphics[width = \linewidth]{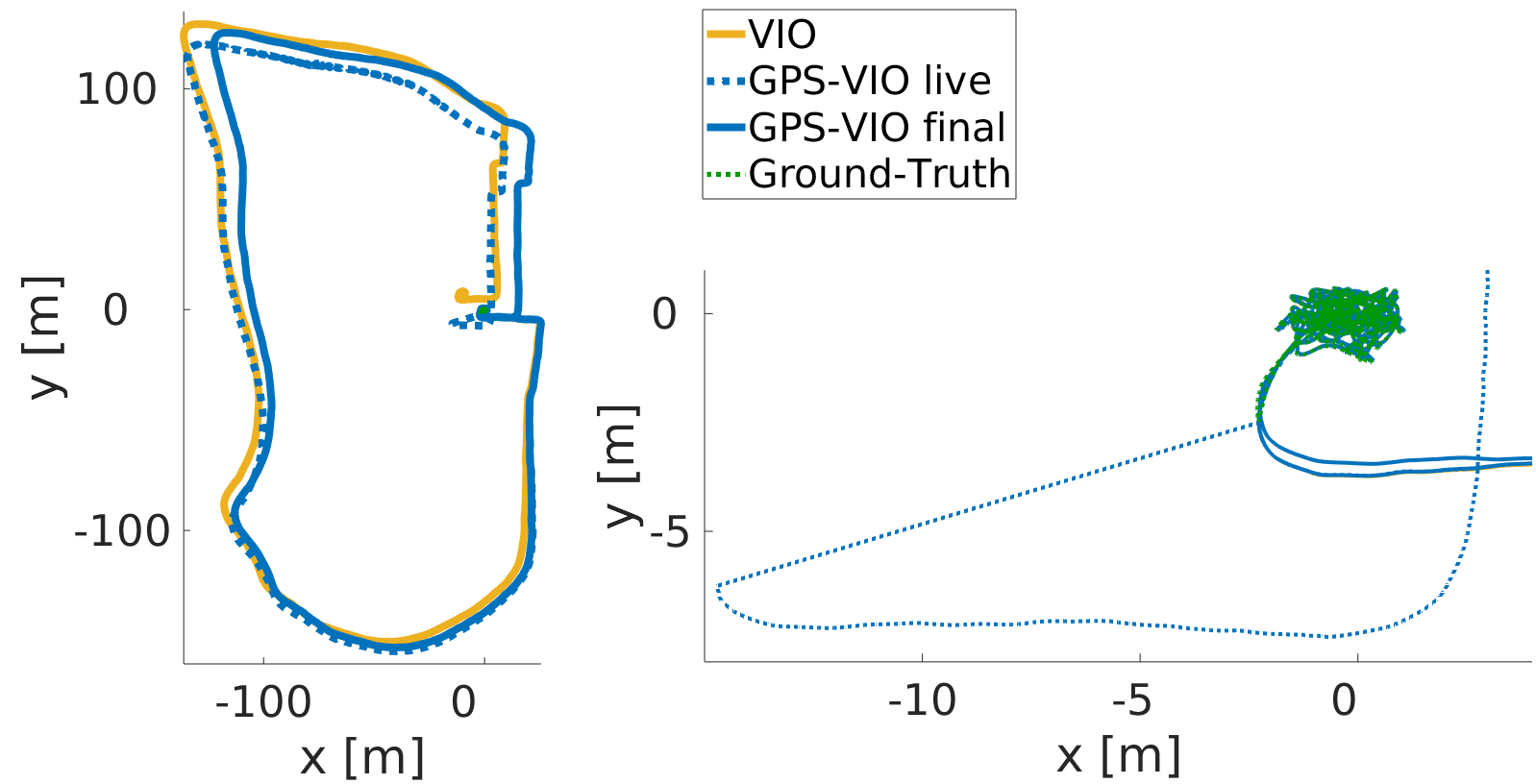}
    \caption{The proposed approach demonstrates its ability to compensate for accumulated long-term drift by GPS loop closure-like alignment strategies. Left: Overall trajectories. Right: zoomed-in view of the live GPS-VIO results in area where measurements become available.}
    \label{fig:tumvi-eval}
\end{figure}
Another benchmark dataset, that is widely known for visual-inertial state estimation approaches, is the TUM-VI benchmark~\cite{TUMVI}. It also provides stereo camera images and high-frequency IMU measurements. Ground-truth data is again provided by a motion capture system, but only for a limited time span in the beginning and at the end of each sequence. As before, we simulate GPS position measurements by disturbing ground-truth positions by some Gaussian noise  with a standard deviation of $\sigma_n = 20$ cm. Using these measurements as input to our GPS-aided VIO, we can simulate the scenario of GPS dropouts over very large periods of time. We demonstrate the global alignment capabilities of the proposed approach on one of the sequences, namely \textit{outdoor4}. Pure VIO is prone to accumulate a large amount of drift due to unavailability of GPS measurements for a large portion of the trajectory. Fig.~\ref{fig:tumvi-eval} shows three different trajectories: 1) the final result of the pure VIO estimator, 2) the live trajectory while running the proposed GPS-aided VIO and 3) the final optimised trajectory running GPS-aided VIO. It is clearly visible that without the fusion of global measurements, the final estimate suffers from a very large drift (around $12$ m final position error). The live trajectory of the GPS-fused approach shows a characteristic jump as soon as measurements after the long GPS dropout trigger a global alignment of the trajectory which immediately corrects the live estimate of the VIO in the global reference frame. The final estimate yields a smooth and globally consistent trajectory. Evaluating the ATE for the segments of the trajectory where ground-truth is available yields an error of only $2.25$ cm.
\begin{figure}[t]
    \centering
    \includegraphics[width=0.87\linewidth]{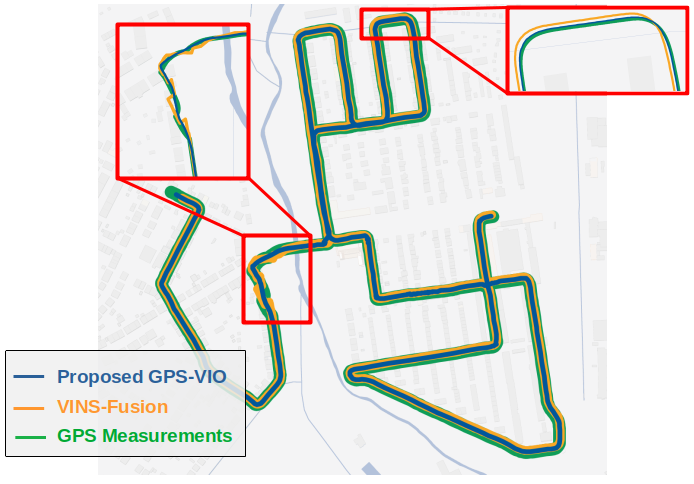}
    \caption{KITTI sequence 09\_30\_drive\_0028: Raw GPS measurements and trajectories estimated by the proposed GPS-VIO approach and VINS-Fusion are visually compared.}
    \label{fig:kitti28}
\end{figure}
\subsection{KITTI Dataset}
So far, the experiments have used only simulated GPS measurements. To verify our system also on realistic data, we run it also on the Raw Kitti Dataset~\cite{KITTI}. Lacking IMU measurements lead to failure of the VI estimator in a lot of the sequences. Hence, we only evaluated our approach on two of the sequences which do not suffer from incomplete measurements (\textit{09\_30\_drive\_0028} and \textit{09\_30\_drive\_0033}). GPS position measurements are provided with uncertainties ranging from $0.02$ to about $1.0$ m. For comparison, also VINS-Fusion has been applied on these sequences. Due to the lack of ground-truth positions, we compute an ATE using the GPS measurements as a reference. On the two evaluated sequences with trajectory lengths of $ 4.23 $ km (\textit{drive\_0028}) and $1.71$ km (\textit{drive\_0033}), we obtain an error of $0.73$ m and $0.08$ m, respectively. These numbers outperform the corresponding $1.04$ m and $0.26$ m errors of VINS-Fusion. We are aware of the limited validity of this evaluation, as the reference positions are also used in state estimation. However, it shows that our resulting estimates are more consistent with the received GPS measurements. This can also be observed visually. Fig.~\ref{fig:kitti28} compares the resulting trajectories on \textit{drive\_028}. In contrast to VINS-Fusion, our proposed approach yields smoother and more consistent results.
%
%
\subsection{Real-World Experiment}
\begin{figure}[t]
    \centering
    \includegraphics[width=0.8\linewidth]{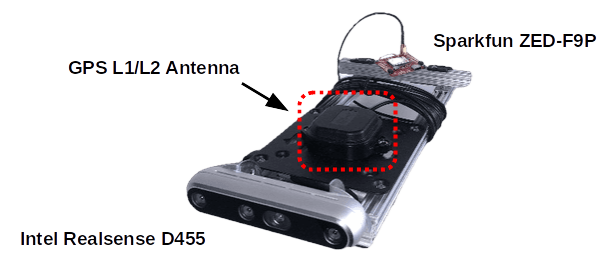}
    \caption{The sensor prototype used for the self-recorded dataset equipped with a RGB-D camera, a GPS sensor and a multiband antenna.}
    \label{fig:prototype}
\end{figure}

Finally, we evaluate the proposed approach also on our own collected real-world dataset. For that reason, we built a prototype sensor rig equipped with a number of sensors shown in Fig.~\ref{fig:prototype}. The overall sensor rig consists of a stereo RGB-D camera, a GPS receiver and a L1/L2 multiband GPS antenna rigidly attached to an aluminium frame. An Intel Realsense D455 captures visual and inertial information. The stereo cameras have been calibrated using the \emph{kalibr} toolbox~\cite{Kalibr}. As a GPS receiver, we used the Sparkfun ZED-F9P which is also RTK (Real Time Kinematic) capable with a localisation error below  $1.4$ cm. The same combination of antenna and receiver is also used to set up a temporary base station. Images are received at a frame rate of 15 Hz while RTK-GPS signals are received at a rate of 10 Hz. IMU measurements are streamed with a frequency of 200 Hz. Both, the base station and the prototype have been wired to a laptop for operation. The devices have been communicating via WiFi. The position of the GPS antenna $\pos{S}{A}$ with respect to the IMU has been measured manually.
\begin{figure*}[htb]
    \centering
    \includegraphics[width = 0.985\linewidth]{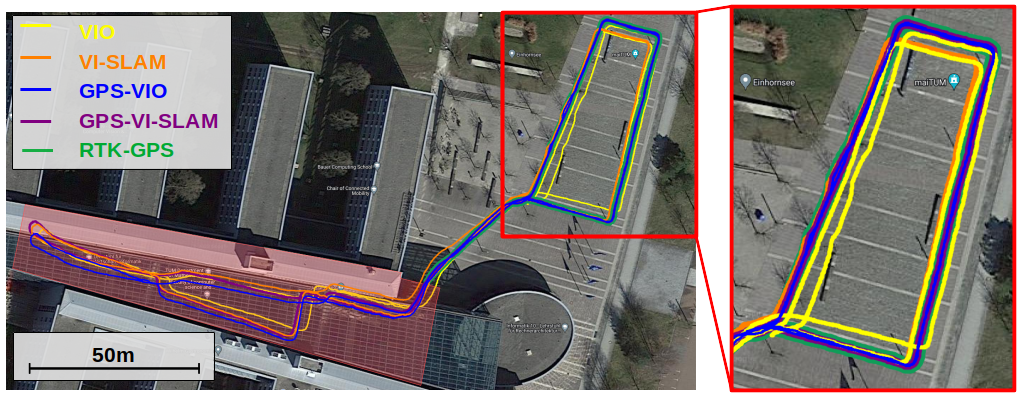}
    \caption{Estimated trajectories for the real-world experiment. OKVIS2 runs both in VIO and in VI-SLAM mode with and without fusing GPS measurements. It can be observed that GPS-aided VIO and VI-SLAM align well with the raw GPS measurements and yield a consistent trajectory loop, whereas pure VI estimates suffer from significant drift. GPS-denied area (inside the building) is highlighted in shaded red.}
    \label{fig:real_world}
\end{figure*}
In the recorded data sequence, we tried to capture a scenario where the proposed global loop closures are triggered due to GPS signals being received again after a long dropout. To enable easy qualitative and visual evaluation, we defined a fixed point as starting and end point of the overall trajectory. Starting from that, we initially went along a loop outside before entering the building. Inside the building, no GPS signals were available any longer and thus, state estimation has to rely on visual-inertial information only. After some time inside the building, we left the building and followed the same loop as before to end our experiment at the starting point.
Estimated final trajectories for GPS-aided VIO and VI-SLAM are visualised next to the results from visual-inertial only estimation in Fig.~\ref{fig:real_world}. The estimator is fused with the RTK-GPS signals to achieve the highest possible accuracy. It can be observed that VIO and VI-SLAM suffer from significant drift. In contrast, GPS-aided estimates align almost perfectly with the measurements where they are available, while still providing more reasonable trajectories indoors (trajectory remains close to existing hallways).
\section{Conclusions}
\label{sec:conclusions}
In this paper, we presented a method combining a VI-SLAM system, based on OKVIS2~\cite{okvis2}, with global position measurement fused in a tightly-coupled approach. The overall VI-SLAM system leverages at the same time stereo visual, inertial and global position measurements together with visual loop closure detection without significant loss of run-time performance compared to~\cite{okvis2}. A novel criterion for the global reference frame initialisation has been introduced that incorporates measurement uncertainties to decide whether the extrinsic transformation between the global and VI reference frame becomes observable. Enabled by the proposed initialisation, as a further new feature, we presented a GPS loop closure-like graph optimisation to eliminate drift that accumulates during long-lasting GPS dropouts. The proposed approach showed its competitiveness to state-of the-art approaches and its robustness to GPS signal outages on public datasets. Finally, its accuracy and globally consistent estimates have also been demonstrated in a real-world setup. 
Future work might include an online calibration of the position of the GPS antenna with respect to the IMU sensor, as well as time offsets. Furthermore, we plan to integrate the algorithm on-board on different mobile robotic platforms enabling applications with indoor-outdoor transitions or operating in forests, urban canyons, etc.\ where GPS is unreliable.



{
\bibliographystyle{IEEEtran}
\bibliography{root}
}

\end{document}